\documentclass[aps,prl,twocolumn,groupedaddress]{revtex4}% PRL
\usepackage{amsmath,amssymb,amsthm,mathrsfs,amsfonts,dsfont} 
\usepackage{graphicx}
\usepackage{subfigure}
\usepackage{amsbsy}
\usepackage{bm}
\usepackage{xcolor}
\usepackage{dcolumn}
\newcommand{\blue}[1]{{\color{black}#1}}
\usepackage{algorithm}
\usepackage{algpseudocode}
\usepackage{amsmath}
\usepackage{array}
\usepackage{booktabs}
\usepackage{physics}

\usepackage{hyperref}% add hypertext capabilities

\def\h{\mathbf{h}}
\def\w{\mathbf{W}}

\begin{document}

\title{Geometric origin of adversarial vulnerability in deep learning}
\author{Yixiong Ren$^{1,2}$}
\thanks{Equal contribution.}
\author{Wenkang Du$^{3}$}
\thanks{Equal contribution.}
\author{Jianhui Zhou$^{1}$}
\email{jhzhou@hmfl.ac.cn}
\author{Haiping Huang$^{3,4}$}
\email{huanghp7@mail.sysu.edu.cn}
\affiliation{$^{1}$Anhui Provincial Key Laboratory of Low-Energy Quantum Materials and Devices,
High Magnetic Field Laboratory, HFIPS, Chinese Academy of Sciences, Hefei, Anhui 230031, People's Republic of China}
\affiliation{$^{2}$University of Science and Technology of China, Hefei 230026, People's Republic of China}
\affiliation{$^{3}$PMI Lab, School of Physics,
Sun Yat-sen University, Guangzhou 510275, People's Republic of China}
\affiliation{$^{4}$Guangdong Provincial Key Laboratory of Magnetoelectric Physics and Devices,
Sun Yat-sen University, Guangzhou 510275, People's Republic of China}
\date{\today}

\begin{abstract}
	Balancing training accuracy and adversarial robustness has beeen a challenge since the birth of deep learning. Here, we introduce a geometry-aware deep learning framework that leverages layer-wise local training to sculpt the internal representations of deep neural networks.
This framework promotes intra-class compactness and inter-class separation in feature space, leading to manifold smoothness and adversarial robustness against white or black box attacks. The performance can be explained by \blue{data-dependent statistical mechanics of integrating out the network parameters}, \blue{supplemented by a phenomenological model} with Hebbian coupling between elements of the hidden representation. Based on the current geometry-aware learning framework, the deep network can assimilate new information into existing knowledge structures 
while reducing representation interference.
\end{abstract}

 \maketitle

%%%%%%%%%%%%%%%%%%%%%%%%%%%%%%%%%%%%%%%%%%%%%%%%%%%%%%%%%%%%%%%%%
%\section{Introduction}
\textit{Introduction.}---
Deep neural networks (DNNs) have achieved remarkable successes across a wide range of applications, especially in scientific discovery~\cite{SAI-2023}, including revealing the brain's mechanisms~\cite{DL-2019}. Recently, DNNs have played a key role in the revolution of natural language processing~\cite{Att-2017,Sparks-2023}. The networks are commonly trained in an end-to-end fashion by backprogation~\cite{BPbrain-2020}, leading to fragile internal representations and associated uncontrolled trade-off between generalization accuracy and adversarial robustness~\cite{Huang-2024}. The trained networks are prone to finding shortcuts (non-conceptual features) to solve the tasks at hand~\cite{Deepbad-2019,Shortcut-2020}.
Therefore, the networks can be easily fooled despite their high test accuracy~\cite{Madry-2018}. The underlying principles behind the accuracy and adversarial robustness remain poorly understood. 

Recent works started to focus on the geometric origin of this trade-off. Empirical studies of the hierarchical nucleation in DNNs were first carried out~\cite{Laio-2020,Donoho-2020,He-2023}, which uncovered how end-to-end training forms a geometric separation of data \blue{but does not reveal the theoretical foundation}. A further conjecture was put forward on the relationship between data concentration and adversarial vulnerability~\cite{Adv-2023}. These works implied that the backpropagation can be replaced by layerwise training with a geometry cost, which was recently realized on a shallow network of one hidden layer~\cite{Xie-2025}. The within-class distance and between-class distance are jointly optimized, leading to a well-controlled trade-off between generalization accuracy and adversarial robustness~\cite{Xie-2025}. However, generalization of this principle to a deep network with an arbitrary number of hidden layers is challenging, as accurate control of intermediate geometry at each hidden layer is required. Therefore, solving the hard-to-balance trade-off in deep networks \blue{is of both practical and theoretical interest}. 

Here, we write the geometry-aware measure into a balance of two terms: the first is designed for a balance between within-class and between-class distances,
expressed as the ratio calibrated to a predefined value (slightly above one); the second is a linear readout of each hidden-layer activity for computational purposes (e.g., classification considered here). \blue{This idea allows a flexible geometric organization of hidden layers, greatly extending our previous work~\cite{Xie-2025} where only a concrete target distance is introduced, rather than the ratio here. Our work thus turns previous empirical observations of hierarchical nucleation~\cite{Laio-2020,Donoho-2020,He-2023} into an algorithmic control of representation geometry.} The learning occurs in a layer-wise fashion, being local \textit{without any global} end-to-end error signal. After the layer-wise training of all hidden layers is completed, a final readout is trained based on the gradually disentangled representations in the deep network. This geometry-aware learning (GAL) thus realizes a controlled disentangling process in deep representation transformation, resembling what occurs in biological neural networks~\cite{Dicarlo-2007,Dicarlo-2012}. The GAL learns semantically meaningful information from noisy data and thus displays strong robustness against adversarial perturbation, which can be explained by \blue{a bottom-up theoretical model and further shares the same broken power-law in neural response of biological vision systems~\cite{Pillow-2025}}, thereby showing a promising angle toward understanding robust learning in both artificial and biological neural networks.

\begin{figure}
\includegraphics[width=0.45\textwidth]{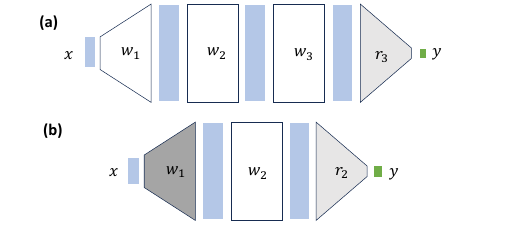} 
 \caption{\label{workflow}%
Illustration of network architecture and layer-wise training. 
(a) The architecture of a neural network with $L=3$ hidden layers and a final readout head $r_3$ trained to predict class probabilities. 
(b) The training scheme for the weight parameters $w_2$ (others are similar). 
In this stage, the parameters $w_1$ have already been trained and are frozen, while $r_2$ denotes the randomly initialized (untrained) readout head tentatively used during the training of $w_2$.}%
\end{figure}

\textit{Geometry-aware deep learning setting.}---
We consider classification tasks and employ an $L$-layer deep fully connected neural network [Fig.~\ref{workflow}(a)]. 
Let $N_l$ denote the dimensionality of the hidden representation $\h_l$ at the layer $l$, and $\w_l$ denote the weight matrix connecting layer $l-1$ to layer $l$. 
The layer-wise transformation is defined as $\h_l = \phi(\w_l^\top \h_{l-1})$, where $\phi(\cdot)$ is a nonlinear activation function, chosen to be tanh in the following.

We adopt a layer-wise training strategy [Fig.~\ref{workflow}(b)], where the network parameters $\w_l$ are optimized in sequence, rather than through end-to-end backpropagation. This makes our learning more biologically plausible~\cite{BPbrain-2020}. 
During training the $l$-th layer ($1 \leq l \leq L$), only the parameters $\w_l$ of this layer are updated, while the parameters of the preceding layers $\w_\ell$ ( $\ell<l$) are frozen. 
The parameters of subsequent layers $\w_{\ell}$ ($\ell>l$) are not involved in the learning of the current layer.

To optimize the parameters $\w_l$ at layer $l$, we design the following local loss function:
\begin{equation}
\mathcal{L}_{\text{local}} =  \mathcal{L}_{\text{GAL}}+\beta \mathcal{L}_{\text{CE}} \label{total_loss},
\end{equation}
where $\mathcal{L}_{\text{CE}}$ is the cross-entropy loss, and $\beta$ is a weighting coefficient. 
During the training of \blue{any hidden layer} $l<L-1$, a frozen linear readout head $r_l$ is attached to the hidden representation $\h_l$ to compute the classification probabilities, which contributes to  $\mathcal{L}_{\text{CE}}$. 
The readout head $r_l$ is randomly initialized from a Gaussian distribution with zero mean and unit variance and \blue{discarded after training the corresponding hidden layer}~\cite{local-2018,Jiang-2021c}.
The first term $\mathcal{L}_{\text{GAL}}$ (layer index $l$ omitted) regularizes the geometric structure of the hidden representation space by promoting intra-class compactness and inter-class separability, explained in detail as follows:
\begin{equation}
\mathcal{L}_{\text{GAL}} = \left| \frac{d_F}{d_B} - \alpha \right|, \label{fbm_loss}
\end{equation}
where $d_{F}$ and $d_{B}$ denote the total pairwise feature distances between samples at layer $l$ from different classes and the same class, respectively:
\begin{subequations}\label{da}
\begin{align}
d_{F} &= \sum_{i,j} [1 - \delta(y_i, y_j)] \cdot \|\h_{l,i} - \h_{l,j}\|_2^2, \\
d_{B} &= \sum_{i,j} \delta(y_i, y_j) \cdot \|\h_{l,i} - \h_{l,j}\|_2^2.
\end{align}
\end{subequations}
Here, $y_i$ is the label of the $i$-th sample, and $\delta(y_i, y_j)$ is the Kronecker delta function, used to select pairs of samples belonging to the same class. 
$\h_{l,i}$ denotes the output feature at layer $l$ of the $i$-th input sample. 
The hyperparameter $\alpha$ controls the desired ratio between inter-class and intra-class (mean) distances. \blue{The value depends on the layer. The current setting applies only to multi-class classification problems, and does not take hierarchically organized hidden states into account. } 

\textit{Data-dependent statistical mechanics.}---
\blue{For an analytic study, we write the absolute penalty [Eq.~\eqref{fbm_loss}] in an equivalent form of squared loss for $P$ training examples of two classes:
\begin{equation}
\tilde{\mathcal{L}}_{\mathrm{local}}
= \sum_{k=1}^{K}\left(d_{F,k}-\alpha d_{B,k}\right)^2
+ \beta \sum_{\mu=1}^{P}\left(y_{\mu}-f_{\mu}\right)^2,
\label{eq:supp-toy-loss}
\end{equation}
where $f_\mu=\mathbf{W}_2^\top\mathbf{h}_\mu$ is the temporary linear readout with quenched weight $\mathbf{W}_2$, and the mean inter-class and intra-class distances are defined respectively:
\begin{subequations}\label{db}
\begin{align}
d_{F,k}&=\frac{1}{\rho (S-1)S}\sum_{\mu,\nu\in\mathcal{B}_k}
\frac{1-y_\mu y_\nu}{2}\|\mathbf{h}_\mu-\mathbf{h}_\nu\|_2^2,\\
d_{B,k}&=\frac{1}{(1-\rho)S(S-1)}\sum_{\mu,\nu\in\mathcal{B}_k}
\frac{1+y_\mu y_\nu}{2}\|\mathbf{h}_\mu-\mathbf{h}_\nu\|_2^2,
\end{align}
\end{subequations}
where $y_\mu=\pm1$, $\mathcal{B}_k$ denotes a mini-batch of size $S$, and the number of inter-class pair is given by $S(S-1)\rho$. A total number of $K$ mini-batches are considered. 
Equation~\eqref{db} coincides with Eq.~\eqref{da} for a special case of two balanced ($\rho=0.5$) classes.}

\blue{Because of layerwise training, we consider the case of one hidden layer of size $N_1$, and thus the Bayesian posterior of weights reads
\begin{align}
p(\mathbf{W}_1|\mathcal{D})
=\frac{1}{Z}
\exp\left[
-\frac{\lambda}{2}\|\mathbf{W}_1\|_F^2
-\frac{1}{T}\tilde{\mathcal{L}}_{\mathrm{local}}(\mathbf{W}_1;\mathcal{D})
\right],
\label{eq:supp-weight-posterior}
\end{align}
where $\mathcal{D}$ denotes the labeled dataset, a Gaussian prior for weights is adopted, $T$ is a temperature, and $Z$ is the partition function. Because of the Gaussian prior, the weight can be integrated out by introducing Dirac-delta functions to constrain the hidden pre-activation $\mathbf{z}^\mu=\frac{1}{\sqrt{N_0}}\mathbf{W}_1^\top\mathbf{x}_\mu$~\cite{Pacelli-2023}, resulting in the following expression:
\begin{align}
Z &=\mathcal{N} \int \prod_{\mu,i} dz^\mu_i\,
\exp\Bigg[
-\frac{1}{2}\sum_{i}\mathbf{z}_i^\top C^{-1}\mathbf{z}_i\nonumber\\
&-\frac{1}{T}\sum_{k=1}^{K}\left(d_{F,k}-\alpha d_{B,k}\right)^2
-\frac{\beta}{T}\sum_{\mu=1}^{P}\left(y_\mu-\mathbf{W}_2^\top\phi(\mathbf{z}^\mu)\right)^2
\Bigg],
\label{eq:supp-latent-posterior}
\end{align}
where $\mathbf{z}_i\in\mathbb{R}^P$, $\mathbf{z}^\mu\in\mathbb{R}^{N_1}$, $C_{\mu\nu}=\frac{\mathbf{x}_\mu^\top\mathbf{x}_\nu}{\lambda N_0}$, $\mathbf{x}_\mu\in\mathbb{R}^{N_0}$, and $\mathcal{N}=\left(\frac{2\pi}{\lambda}\right)^{N_0N_1/2}\left[(2\pi)^P{\rm det}\,C\right]^{-N_1/2}$. The test error $\langle\epsilon_{\mathrm{g}}(\mathbf{x}_0,y_0)\rangle$ can be derived from the partition function by adding a source term, and the mean distance $\langle\|\phi(\mathbf{z}_\mu)-\phi(\mathbf{z}_\nu)\|^2\rangle$ can also be computed using the optimal hidden state derived from Eq.~\eqref{eq:supp-latent-posterior}. Details are presented in the SM~\cite{SM}. For simplicity, we focus on $P=S$ and $K=1$ in our theoretical analysis.}

\begin{figure}
    \centering
    \includegraphics[width=0.5\textwidth]{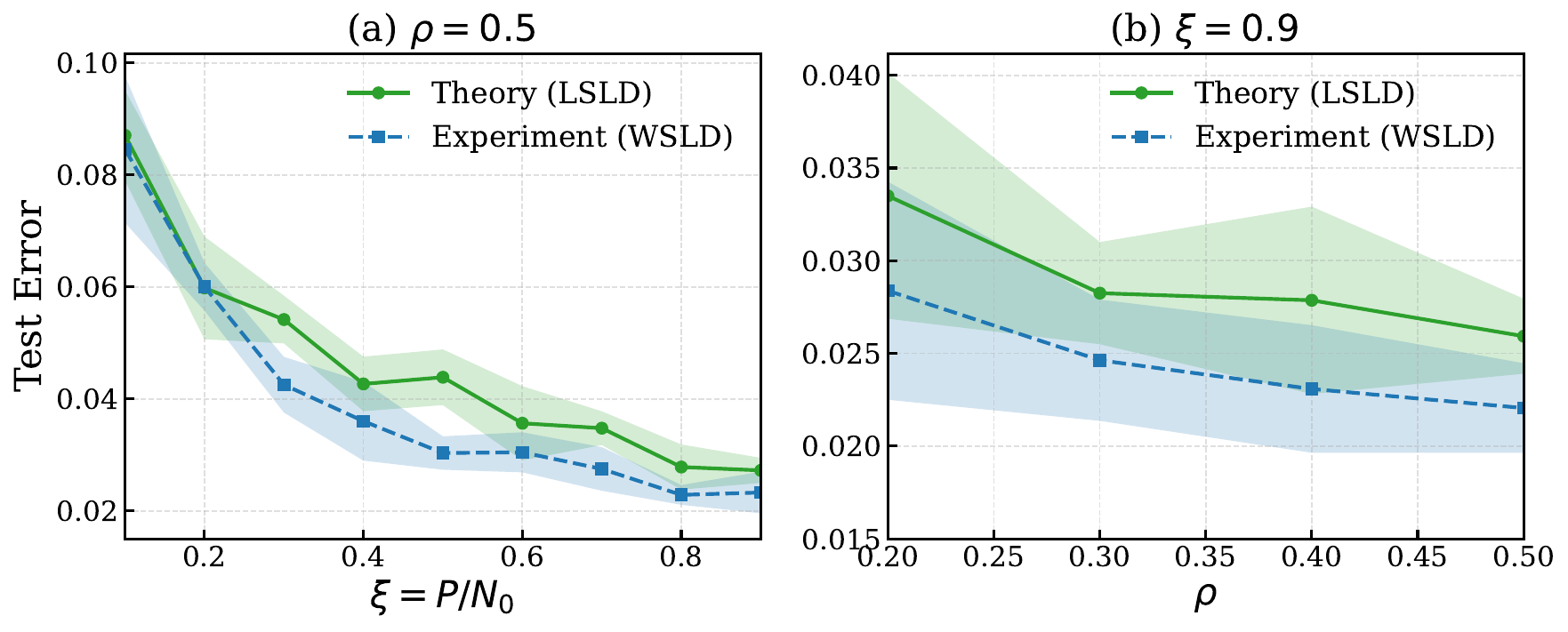}
    \caption{\blue{Test error comparison between theory and experiment across data sizes and class balance ratios.
   A single-hidden-layer neural network with an input dimension of $N_0 = 784$ and hidden-layer width $N_1 = 100$ is used to learn MNIST dataset (digits 0 and 1), with the target distance ratio $\alpha = 1.1$, loss weighting coefficient $\beta = 1.0$, temperature $T = 0.01$, and regularization coefficient $\lambda = 1.0$. 
    (a) Test error vs. data density ($\xi$) under a perfectly balanced class condition ($\rho = 0.5$). The theoretical prediction derived from the Latent Space Langevin Dynamics (Theory LSLD~\cite{SM}, green solid line) and the experimental Weight Space Langevin Dynamics (Experiment WSLD, blue dashed line) are compared. Shaded regions represent the standard deviation across multiple independent runs. 
    (b) Test error vs. balance ratio $\rho$ at a fixed $\xi = 0.9$.} }
    \label{fig2}
\end{figure}

\blue{This data-dependent statistical mechanics formula transforms a weight-description (learning) into a representation-description, where the propagator $C$ plays a key role. We can therefore study how the internal representation varies with respect to training data size, balance ratio, and adversarial attacks. First, the test error decreases with the data size in Fig.~\ref{fig2} (a), while Fig.~\ref{fig2} (b) confirms that the test performance improves with the balance ratio $\rho$, which is expected as more inter-class interaction information facilitates cluster separation and is also consistent with training results in the multiple-layer setting~\cite{SM}. In addition, our theoretical prediction of distance $d_F/d_B$ is compatible with experiments, slightly above the target ratio (Fig.~\ref{fig2}). The two-dimensional visualization of the latent space shows that the hidden geometry is indeed well-controlled, and even in the presence of an adversarial attack (an $\ell_{\infty}$-norm bounded fast gradient sign method (FGSM) attack using trained network parameters~\cite{Szegedy-2014,Good-2015}), the clustering geometry still holds, although the class boundary becomes crowded. We also build a phenomenological model of GAL as a supplementary description, using Hebbian learning, supporting the pattern-condensation phenomenon and sphere-like topology (see~\cite{SM}). }

\begin{figure}
    \centering
    \includegraphics[width=0.5\textwidth]{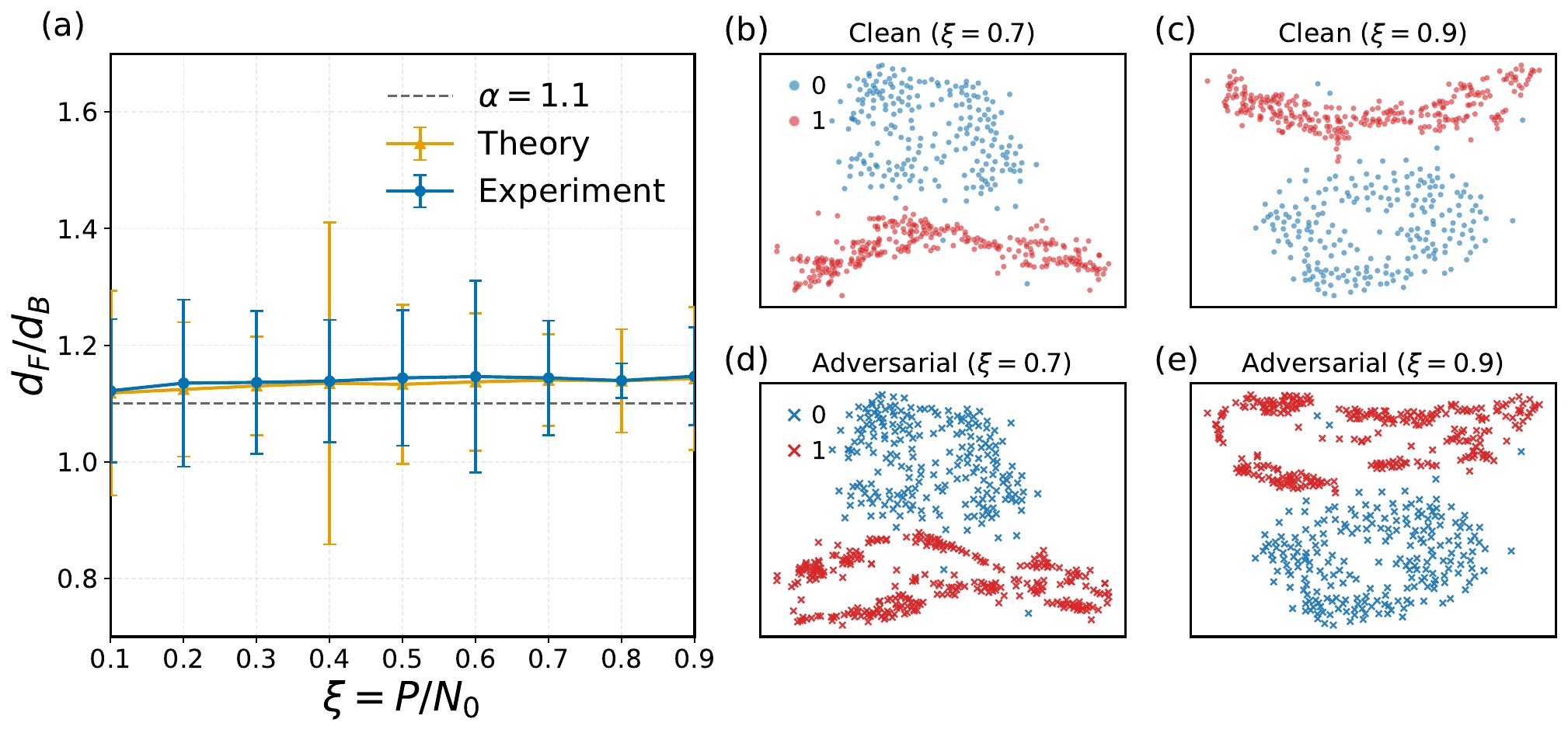}
    \caption{\blue{Geometry of latent space.
    (a) Distance ratio $d_F/d_B$ vs. data density. Target value $\alpha = 1.1$ (gray dashed line). 
    (b)-(c) Clean-data latent manifolds. t-SNE visualizations are shown.  
    (d)-(e) Adversarial-data latent manifold. An $\ell_\infty$-norm attack with strength $\varepsilon = 0.15$ is used. }}
    \label{fig3}
\end{figure}

 \textit{Results and discussion.}---
 In the following, we consider the network
architecture $784$-$1000$-$1000$-$1000$-$10$ trained with a full dataset of MNIST or CIFAR-10 that is greyscaled and downscaled to $28 \times 28$, and pixels of both images are normalized to the range $[-1, 1]$. 
\blue{Other datasets and architectures are reported in~\cite{SM}}.

\begin{figure}
    \includegraphics[width=0.5\textwidth]{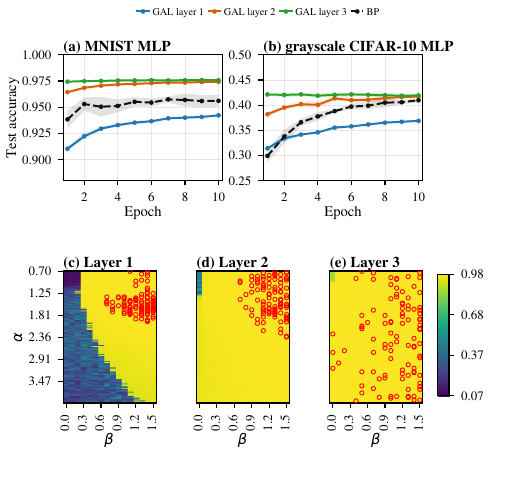} 
     \caption{\label{data_acc}%
  Classification accuracy on the test dataset across network layers. 
    (a) Accuracy on the MNIST dataset; 
    (b) Accuracy on the grayscale-transformed CIFAR-10 dataset. 
    Each curve corresponds to a different hidden layer, with shaded regions indicating standard deviation across ten independent runs. 
    (c-e) Effects of hyperparameters $(\alpha,\beta)$ on the network performance for each hidden layer (from shallow to deep). Circles represent the top $100$ accuracies.
    }\end{figure}
    
   As shown in Fig.~\ref{data_acc}, GAL reaches a similar accuracy with the backpropagation~\cite{Li-2020,Li-2023}. 
Notably, deeper layers achieve a higher accuracy, reflecting their enhanced representational capacity and ability to capture more abstract features (detailed below). 
 How hyperparameters $(\alpha,\beta)$ affect the accuracy is illustrated in Fig.~\ref{data_acc} (c-e). A t-SNE~\cite{Hinton-2008} visualization of hidden representations across layers is shown in the SM~\cite{SM} and confirms that
  the geometry penalty drives learning to facilitate generalization (within the same class yet with certain dispersion) and discrimination (from different classes). Both generalization and discrimination can be controlled layer by layer, ensuring a clear and robust decision boundary between different categories of input images \blue{(prototype centroids maintain an approximately orthogonal relationship~\cite{SM})}. This establishes the foundation for the following property of robustness against adversarial attacks. 

\begin{figure}
    \includegraphics[width=0.50\textwidth]{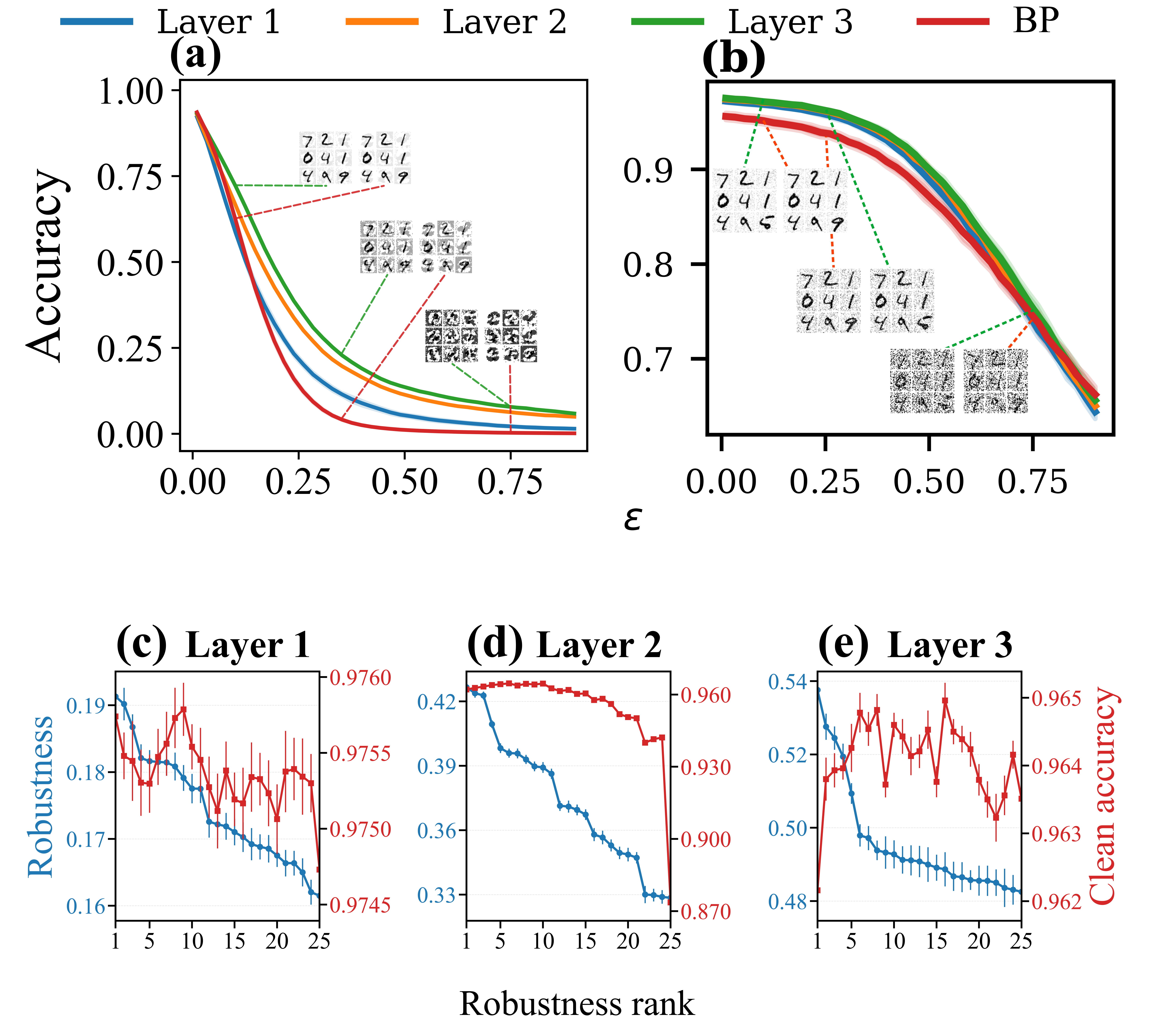} 
     \caption{\label{robust}%
    Network robustness analysis under different types of adversarial attacks. Results are averaged over ten independent runs.
   (a) Test accuracy under FGSM attacks. (b) Test accuracy under Gaussian noise attacks. 
    The horizontal axis denotes the attack strength $\varepsilon$, and the vertical axis indicates the test accuracy after the attack. 
     A multi-layered perceptron (MLP) trained with end-to-end backpropagation is also compared. \blue{(c-e) Layer-dependent adversarial robustness as a function of clean-data accuracy. The adversarial robustness is measured by
     the area under the accuracy-$\varepsilon$ curves. The hyperparameters $(\alpha,\beta)$ are taken from the top twenty-five clean-data accuracy values (horizontal axis) in Fig.~\ref{data_acc}.}
    }\end{figure}
    
The standard deep networks trained with backpropagation were found to be easily fooled by adversarial examples~\cite{Szegedy-2014,Good-2015}. Adversarial examples
refer to the inputs corrupted by tiny variations (at least imperceptible to humans) that dramatically change the network output (e.g., misclassification with high confidence), which poses a significant
challenge to the practical applications of deep networks (e.g., confusion of traffic signs)~\cite{Deepbad-2019}. To show how the proposed GAL can mitigate the adversarial attack,
we consider the FGSM~\cite{Szegedy-2014,Good-2015,Jiang-2021c,Xie-2025} and additive white noise attacks to the inputs of trained neural networks~\cite{Jiang-2021c,Xie-2025}. 

As expected, the accuracy of all models decreases with increasing attack strength. 
However, deeper layers exhibit significantly stronger robustness, with slower performance degradation under both attack types [Fig.~\ref{robust} (a,b)]. This observation is consistent with
our previous theoretical analysis. \blue{Note that backpropagation is not robust. While the robustness can be improved by adding adversarial examples for training (adversarial training~\cite{Madry-2018}, see also~\cite{SM}), an extra optimization (min-max) should be done, and moreover, the adversarial training renders the standard accuracy on clean data to drop~\cite{Adv-2018d}, especially on complex datasets such as CIFAR-10 and ImageNet.
Fig.~\ref{robust} (c-e) highlights the advantage of our layer-wise geometry-aware deep learning in terms of balancing test accuracy and adversarial robustness. }

It was revealed that recorded population activity in the visual cortex of awake mice shows power-law behavior in the principal component spectrum of the population responses~\cite{Nature-2019}, i.e., the $n$th biggest principal component (PC) variance scales as $n^{-\gamma}$, where $\gamma$ is the exponent of the power law. A larger exponent value (than one typically) reflects an intrinsic property of a smooth coding in biological neural networks~\cite{Nature-2019}. Our GAL bears a certain level of biological plausibility, and thus, one natural question is how smooth the representation manifold is in our case.  

\begin{figure}
    \includegraphics[width=0.50\textwidth]{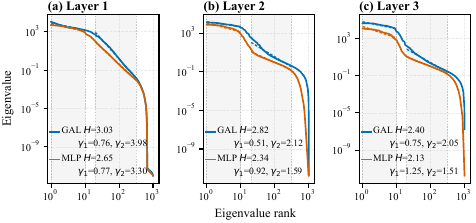} 
     \caption{\label{eigenspectrum}%
   \blue{ Log-log scale linear fit on eigenspectra of feature covariance matrices for each hidden layer in the trained network, compared with backpropagation. 
    The horizontal axis denotes the eigenvalue ranking (sorted in descending order), and the vertical axis shows the corresponding eigenvalue magnitude. 
    Each solid line represents the full eigenspectrum of one layer, while the dashed line indicates a linear fit. 
    The estimated spectral exponents $\gamma$ and the entropy are reported in the legend. }
    }\end{figure}

We analyze the eigenspectra of the feature covariance matrices across the three hidden layers of the trained network in Fig.~\ref{eigenspectrum} and find that they all exhibit power-law decay yet with two groups of exponents (broken power law). \blue{Compared to GAL, the exponents of the first power law increase across layers in the end-to-end training, while those of the second power law imply that the spectrum decays more slowly than GAL, explaining the weak robustness of end-to-end training. A flat eigenspectrum can enhance the capacity of neural codes, while smoothness, related to a fast decay of the spectrum,
determines the generalization capability~\cite{Nature-2019,Pillow-2025}, i.e., in the converse (non-smooth) case, nearby stimuli elicit widely distributed responses and are thus fragile to perturbations. The eigenvalues can be normalized to yield a fraction $p_i=\frac{\lambda_i}{\sum_j\lambda_j}$ ($\lambda_i$ is non-negative), which yields further the entropy $H=-\sum_i p_i\ln p_i$. A high value of the entropy implies high-dimensional isotropic representation, while a low value indicates a low-dimensional anisotropic representation. A progressive decrease of $H$ suggests a low-dimensional semantic manifold emerges from layered learning, which accounts for the robust classification.}  This property of GAL is consistent with a previous empirical study of local learning~\cite{Jiang-2021c} \blue{and is also recently observed as a broken power-law in mouse primary visual cortex~\cite{Pillow-2025}}. \blue{This property holds in wider networks, but the broken power-law behavior is altered when the spatial structure of the sensory input is destroyed despite retaining the same statistics (see SM~\cite{SM}).} 

\textit{Conclusion and outlook.}---
In this work, we address the geometric origin of adversarial vulnerability and propose a layer-wise geometry-aware training strategy for deep learning, challenging the current end-to-end backpropagation in the following three aspects. 
First, the representation at each layer is trained independently with a local random classifier and geometric constraints on the hidden representation at that layer, and thus the learning is local without the need to store all intermediate activities and weight symmetry to propagate the global error. Second, the geometric property of the hidden representation can be well controlled by a single hyperparameter, the ratio between expansion and contraction, \blue{conceptually explained by data-dependent statistical mechanics transforming the Bayesian posterior weight-description into representation-description.} Third, this geometry-aware learning leads to smooth and differentiable manifolds, \blue{thereby balancing the discrimination and generalization}. Thanks to these three intriguing properties, the current work would further provide a guideline for better understanding how artificial and biological neural networks align at the algorithmic level, \blue{and for establishing learning mechanics to interpret how geometry and weight dynamics are combined to yield a robust learning system}.

%%%%%%%%%%%%%%%%%%%%%%%%%%%%%%%%%%%%%%%%%%%%%%%%%%%%%%%
\section*{Acknowledgments}
This research was supported by the National Natural Science Foundation of China for
Grant number 12475045 (H.H.), and National Key R$\&$D Program of the MOST of China under Grant No. 2024YFA1611300 (J.Z.), and the National Natural Science Foundation of China under Grants No. 12174394 (J.Z.), and the HFIPS Director’s Fund under Grants No. BJPY2023B05 (J.Z.), and Anhui Provincial Major S$\&$T Project (s202305a12020005) (J.Z.), and the Basic Research Program of the Chinese Academy of Sciences Based on Major Scientific Infrastructures (Grant No. JZHKYPT-2021-08) (J.Z.). and Guangdong Provincial Key Laboratory of Magnetoelectric Physics and Devices (No. 2022B1212010008) (H.H.), and Guangdong Basic and Applied Basic Research Foundation (Grant No. 2023B1515040023) (H.H.).
%%%%%%%%%%%%%%%%%%%%%%%%%%%%%%%%%%%%%%%%%%%%%%%%%%%%%%%%%%%%%%%

Codes are available in our Github:~\href{https://github.com/RenYixiong-ai/GAL}{https://github.com/RenYixiong-ai/GAL}.

%\newpage
%\onecolumngrid
%\begin{widetext}
%\appendix

%\end{widetext}

%%%%%%%%%%%%%%%%%%%%%%%%%%%%%%%%%%%%%%%%%%%%%%%%%%%%%%%%%%%%%%%%%%%%%
%\bibliography{ref}

%%%%%%%%%%%%%%%%%%%%%%%%%%%%%%%%%%%%%%
\end{document}